\documentclass[10pt,twocolumn,letterpaper]{article}

\usepackage{3dv}
\usepackage{times}
\usepackage{epsfig}
\usepackage{graphicx}
\usepackage{amsmath}
\usepackage{amssymb}
\usepackage{color}

\usepackage{booktabs}
\usepackage{siunitx}
\usepackage{pifont}

\usepackage[pagebackref=true,breaklinks=true,colorlinks,bookmarks=false]{hyperref}

\threedvfinalcopy %

\newcommand*\colourcheck[1]{%
  \expandafter\newcommand\csname #1check\endcsname{\textcolor{#1}{\ding{52}}}%
}
\newcommand*\colourcross[1]{%
  \expandafter\newcommand\csname #1cross\endcsname{\textcolor{#1}{\ding{56}}}%
}

\newcommand{\dsecflow}{DSEC-Flow}

\colourcheck{green}
\colourcheck{yellow}
\colourcheck{orange}
\colourcross{red}

\definecolor{somegray}{rgb}{0.5, 0.5, 0.5}
\newcommand{\darkgrayed}[1]{\textcolor{somegray}{#1}}
\makeatletter
\newcommand*\titleheader[1]{\gdef\@titleheader{#1}}
\AtBeginDocument{%
  \let\st@red@title\@title
  \def\@title{%
    \vskip-3em
    \bgroup\normalfont\large\centering\@titleheader\par\egroup
    \vskip1.5em\st@red@title}
}
\makeatother

\titleheader{\darkgrayed{This paper has been accepted for publication at the\\
International Conference on 3D Vision (3DV), 2021.
\copyright IEEE}}

\def\threedvPaperID{323} %

\title{E-RAFT: Dense Optical Flow from Event Cameras}

\ifthreedvfinal\fi
\begin{document}

\author{Mathias Gehrig\thanks{Equal contribution}\qquad Mario Millh\"ausler$^{*}$\qquad Daniel Gehrig\qquad Davide Scaramuzza\\
Dept. Informatics, Univ. of Zurich and \\
Dept. of Neuroinformatics, Univ. of Zurich and ETH Zurich\\
}

\maketitle
\thispagestyle{empty}

\begin{abstract}
   We propose to incorporate feature correlation and sequential processing into dense optical flow estimation from event cameras.
   Modern frame-based optical flow methods heavily rely on matching costs computed from feature correlation.
   In contrast, there exists no optical flow method for event cameras that explicitly computes matching costs.
   Instead, learning-based approaches using events usually resort to the U-Net architecture to estimate optical flow sparsely.
   Our key finding is that the introduction of correlation features significantly improves results compared to previous methods that solely rely on convolution layers.
   Compared to the state-of-the-art, our proposed approach computes dense optical flow and reduces the end-point error by 23\% on MVSEC.
   Furthermore, we show that all existing optical flow methods developed so far for event cameras have been evaluated on datasets with very small displacement fields with maximum flow magnitude of 10 pixels.
   Based on this observation, we introduce a new real-world dataset that exhibits displacement fields with magnitudes up to 210 pixels and 3 times higher camera resolution.
   Our proposed approach reduces the end-point error on this dataset by 66\%.
\end{abstract}

\section*{Multimedia Material}
{
\noindent Code: \url{https://github.com/uzh-rpg/E-RAFT}\\
Dataset: \url{https://dsec.ifi.uzh.ch}\\
Video: \url{https://youtu.be/dN8fl7-XfNw}
}

\section{Introduction}

\begin{figure}[ht]
    \centering
    \includegraphics[width=0.49\linewidth]{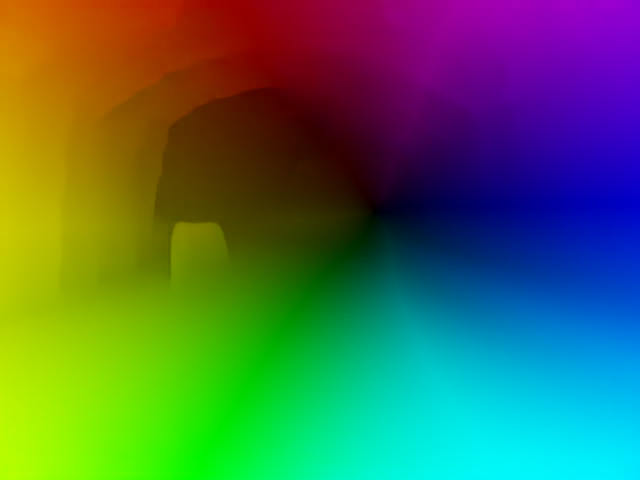}
    \includegraphics[width=0.49\linewidth]{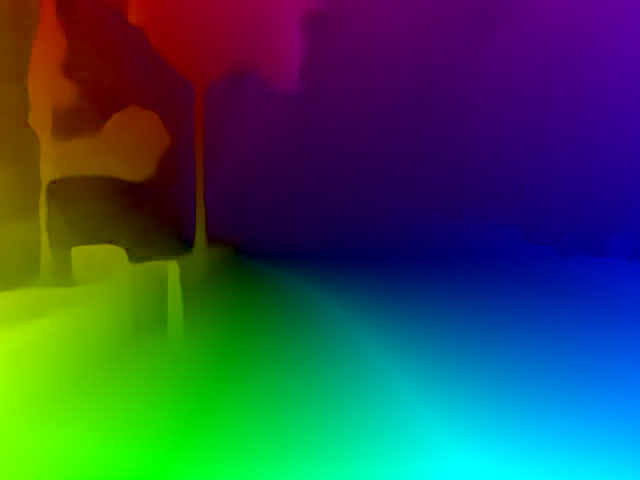}\\
    \includegraphics[width=0.49\linewidth]{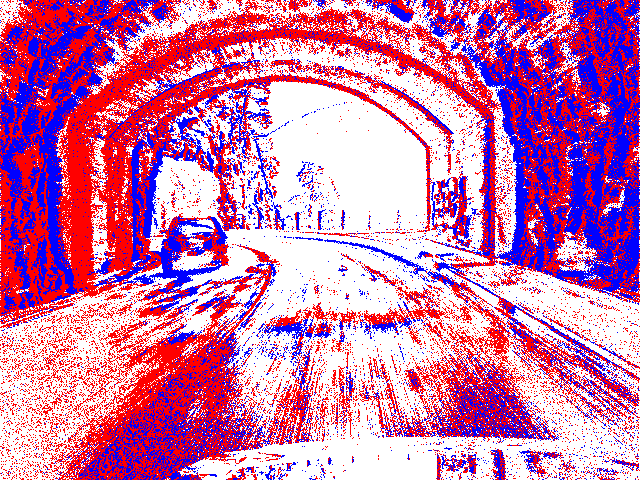}
    \includegraphics[width=0.49\linewidth]{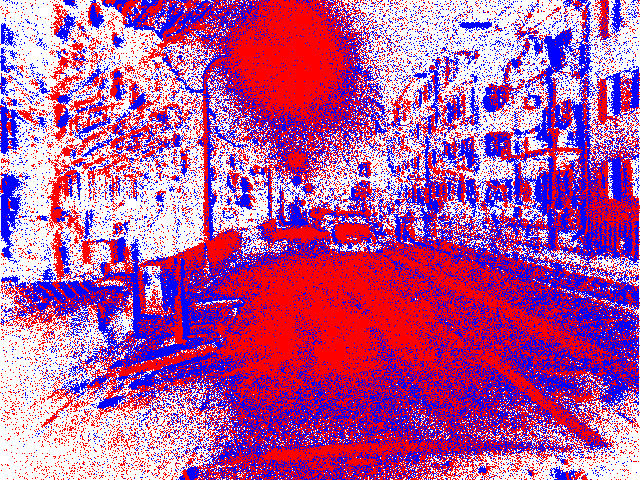}\\
    \includegraphics[width=0.49\linewidth]{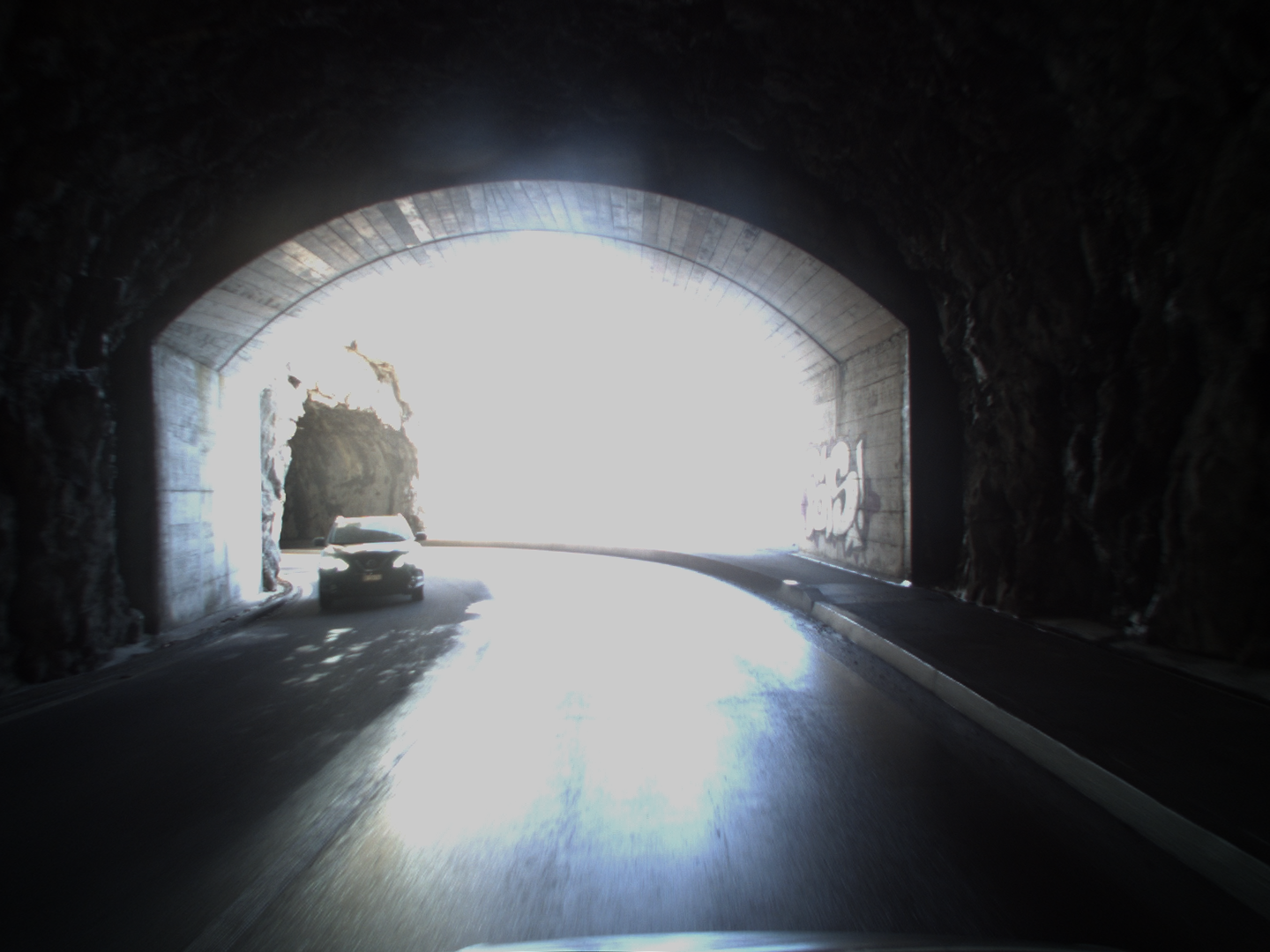}
    \includegraphics[width=0.49\linewidth]{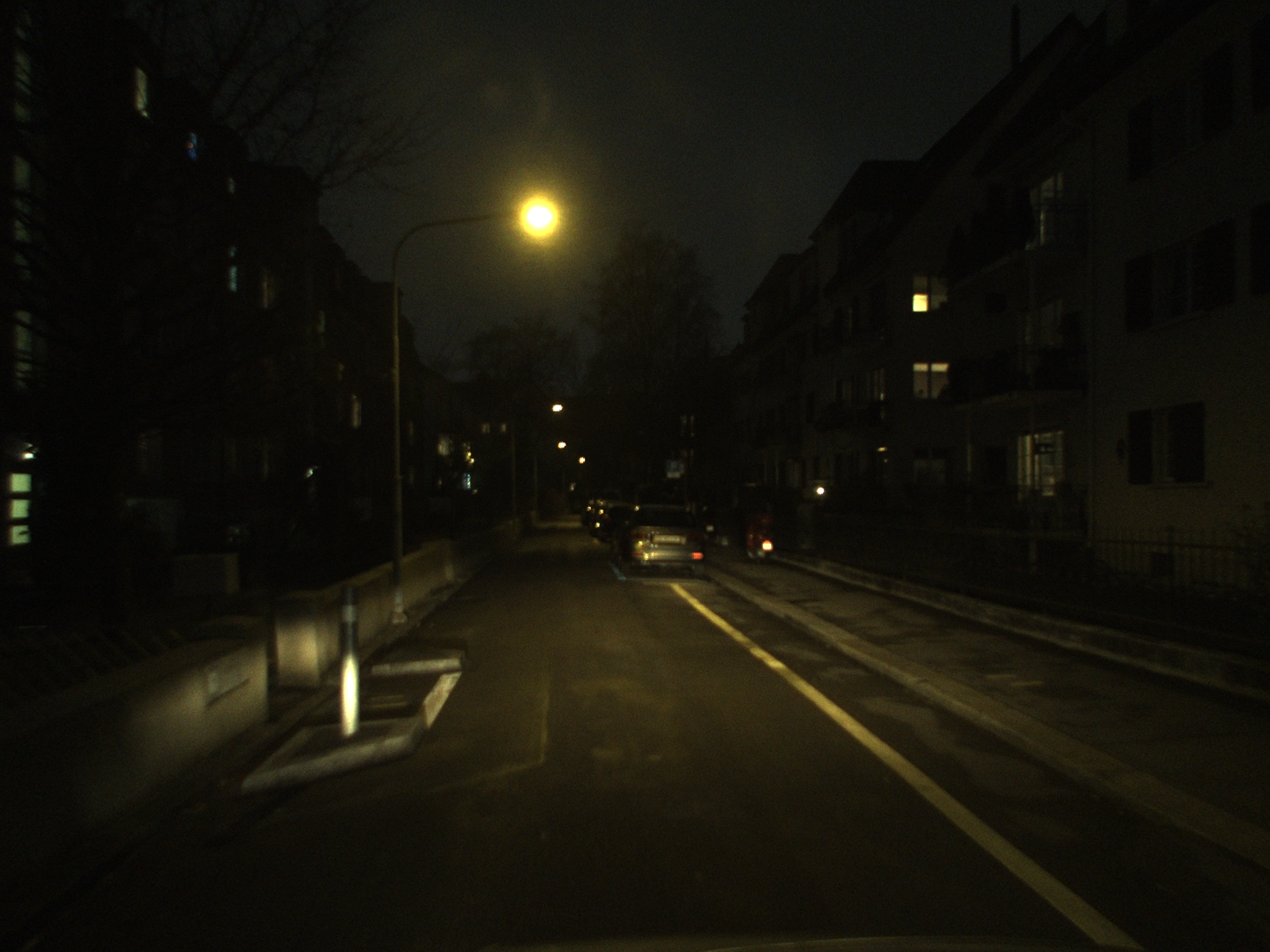}
    \caption{Optical flow predictions from our proposed approach trained on our novel optical flow dataset for event-based vision. The second and third row visualize input event data and images of the same scene respectively.}
    \label{fig:eyecatcher}
\end{figure}
Optical flow estimation is a fundamental computer vision task that informs about movement in image space. It serves as a fundamental building block for visual odometry \cite{qin2018vins} for robotics and AR/VR, autonomous driving \cite{janai2020computer}, action recognition \cite{Karen14NeurIPS}, high dynamic range (HDR) imaging \cite{kalantari2017deep}, and computational videography \cite{niklaus20cvpr}. Traditionally, this problem is formulated as finding pixel correspondences between two images.

While earlier top-performing methods were mostly based on energy minimization techniques \cite{Horn81ai}, current state of the art is dominated by neural network-based approaches \cite{teed20eccv}.

Decades of research on image-based optical flow estimation have resulted in impressive results on public benchmarks \cite{Menze2015CVPR}.
However, image-based methods struggle in presence of motion blur \cite{Butler:ECCV:2012} or over-saturated image regions due to limited dynamic range of image sensors.
A promising sensor for optical flow estimation is the event camera.

Event cameras drastically differ from image-based sensors in their way of acquiring data. In contrast to frame-based cameras, which capture images at regular intervals, event cameras report per-pixel brightness changes as a stream of asynchronous events. For a detailed survey on event cameras, we refer to \cite{Gallego20pami}. The distinct advantages of event cameras are high temporal resolution (microseconds), high dynamic range ($>120$ dB), and no motion blur. All three aspects are essential for high-quality optical flow estimation in real-world applications. However, these advantages are tied to the fundamentally different data format resulting from event cameras.

Effectively extracting motion from an asynchronous event stream is a non-trivial task.
Unlike images, which can be directly used as input to convolutional neural networks, event data is sparse, irregular, and asynchronous in nature.
Prior work on optical flow estimation from events proposed spatio-temporal plane-fitting \cite{Benosman14tnnls}, variational optimization based on discretized spatio-temporal volumes \cite{Bardow16cvpr} and, more recently, convolutional architectures on preprocessed event data \cite{Zhu18rss, Zhu19cvpr, Ye19arxiv}.
Similarly to image-based optical flow, the current state of the art is based on data-driven neural network approaches.
These data-driven methods have in common that they are only evaluated for \emph{sparse} optical flow prediction while relying on a U-Net architecture \cite{ronneberger2015u} to extract motion features from events converted into a tensor representation. 

In contrast to event-based optical flow methods, modern image-based optical flow approaches make use of cost volumes to extract discriminant features \cite{Xu_2017_CVPR, Sun18cvpr, yang2019volumetric, teed20eccv}. Typically, a neural network computes image features at a lower resolution, which are then used to compute feature correlations between the source and the target descriptors. The feature correlations are summarized in a cost volume that is processed with CNN layers \cite{Sun18cvpr, yang2019volumetric}. Teed et al. \cite{teed20eccv} chose instead to extract features from the cost volume via an iterative update scheme.

In contrast to prior work, we focus on \emph{dense} optical flow estimation for event cameras and utilize cost volumes instead of using the U-Net architecture.
First, we show that a state-of-the-art method designed for images can be adopted to work exceptionally well on event-data, outperforming all previously proposed methods consistently.
Then, we exploit the fact that event data is sequential and propose a differentiable warm-starting technique that initializes the flow estimate for the next time step. 
Differently from previous work \cite{teed20eccv}, we perform backprogagation through time via the differentiable warm-starting module.
Our experiments suggest that incorporating this warm-starting technique in the training pipeline improves the performance on sequential event data further.
More importantly, our proposed approach can be directly used for fine-tuning our non-recurrent model for sequential data because it does not introduce any additional parameters.

Until now, data driven optical flow methods on events have solely been evaluated on the MVSEC dataset \cite{Zhu18rss}.
Our analysis suggests that the difficulty of this dataset is very limited.
For example, 95\% of ground truth flow on the outdoor evaluation sequence \cite{Zhu18rss} has a magnitude of less than 3 pixels (see figure \ref{fig:flow_hist}).
Based on this observation, we extend a recently proposed real-world driving dataset \cite{gehrig2021dsec} with optical flow ground truth. This dataset features a maximum flow magnitude of up to 210 pixels and an event camera resolution that is three times higher than MVSEC.
We train and evaluate a state-of-the-art event-based method on this dataset and show that our proposed approach outperforms existing approaches that have so far only been tested on MVSEC by 23\% EPE (0.27 vs 0.35) and by 66\% EPE (0.79 vs 2.32) on our proposed dataset.

In summary, our contributions are

\begin{itemize}
    \item The first optical flow approach for event cameras that is designed and evaluated for dense optical flow estimation.
    \item The first optical flow method for event cameras, which utilizes cost volumes and introduces recurrency to incorporate temporal priors. Previous event-based optical flow methods rely on plane fitting, variational optimization or pure CNN layers to estimate optical flow.
    \item A novel real-world optical flow dataset for event-based vision with significantly larger pixel displacements, higher resolution than current datasets, and higher ground truth quality.
    \item Controlled experiments and evaluation on both MVSEC and our proposed dataset that suggest that both cost volumes and temporal recurrency are powerful concepts for event-based optical flow estimation outperforming purely convolutional architectures.
\end{itemize}

\section{Related Work}\label{sec:relwork}
\subsection{Learning-based Optical Flow Estimation on Images}

Data-driven approaches for optical flow estimation has been pioneered by FlowNet \cite{Dosovitskiy15iccv}.
In the following years, optical flow estimation has witnessed great progress through principled neural network design \cite{zhao2020maskflownet, hui2018liteflownet, hur2019iterative, liu2019selflow, Sun18cvpr, sun2019models, yang2019volumetric, teed20eccv}. 
As a result a few core ingredients for high performance optical flow estimation have emerged: \emph{(i)} coarse-to-fine processing, \emph{(ii)} the use of feature correlation volumes, and \emph{(iii)} feature warping \cite{zhao2020maskflownet, hui2018liteflownet, hur2019iterative, liu2019selflow, Sun18cvpr, sun2019models, yang2019volumetric}.
Teed et al. \cite{teed20eccv} builds on these ideas and extends them by constructing correlation volumes at multiple scales and introducing an iterative approach to extract optical flow from them, reaching state-of-the-art results on the Sintel\cite{Butler:ECCV:2012} and KITTI\cite{Geiger13ijrr} benchmarks.

While these methods work well in standard conditions, their performance is intricately tied to the quality of the images used. 
These algorithms are thus not able to cope with image degradations, that occurr in challenging scenarios, such as in HDR scenes and during high-speed motion.   

In this work, we seek to address this intrinsic limitation of current approaches, by proposing the use of event cameras \cite{Lichtsteiner08ssc} which do not suffer from motion blur or limited dynamic range.

\subsection{Event-based Optical Flow}\label{sec:relw_event_flow}

Algorithmic methods for event-based optical flow estimation can be summarized in three categories. First, iterative asynchronous methods inspired by \cite{benosman2012asynchronous, Gehrig19ijcv} by the well-known Lucas-Kanade algorithm \cite{Lucas81ijcai}.
The latter also leverages image data for photometric feature tracking.
Second, plane fitting methods \cite{brosch2015event, Mueggler15icra} that exploit the local plane-like shape of spatio-temporal event streams.
Third, variational optimization based approaches \cite{Bardow16cvpr, pan2020single} that incorporate image data \cite{pan2020single} or simultaneously estimate image intensity \cite{Bardow16cvpr}.
These methods directly work on event data and often incorporate strong assumptions, such as brightness constancy \cite{Bardow16cvpr, pan2020single}, into their models. In contrast, our approach learns abstractions and priors from data and does not rely on such assumptions.

Another line of research pursues learning-based approaches and exclusively adopts the self-supervised learning framework \cite{Zhu19cvpr, lee2020spike, Ye21iros, paredes2020back, Zhu18rss}.
Supervision is either provided by images \cite{Zhu18rss, Zhu19cvpr,lee2020spike} or events only \cite{Ye21iros,paredes2020back}.
However, these methods, except for \cite{Ye21iros}, choose to mask their optical flow prediction for evaluation due to limiting accuracy in regions where no events are present. On the other hand, Ye et al. \cite{Ye21iros} assume that the scene is static and predict optical flow via depth and ego-motion.
All aforementioned approaches have in common that they use the U-Net architecture to directly estimate and evaluate \emph{sparse} optical flow from event representations.
In contrast to above methods, we predict and evaluate \emph{dense} optical flow, make explicit use of longer temporal history, and leverage correlation features from cost volumes to estimate pixel correspondences for large displacements.
Finally, we depart from the common theme of using a U-Net architecture for event-based optical flow estimation while at the same time outperforming existing methods consistently on multiple datasets.

\subsection{Datasets for Event-based Optical Flow}\label{sec:rel_datasets}
\begin{table*}[ht]
    \centering
    \begin{tabular}{@{}l|cclcc@{}}
\toprule
\textbf{Dataset} & \multicolumn{1}{l}{Train/Test Split} & \multicolumn{1}{l}{Groundtruth Quality} & Resolution (MP) & \multicolumn{1}{l}{Translational Motion} & \multicolumn{1}{l}{Outdoor} \\ \midrule
DVSFLOW16        & \redcross                                    & \textbf{high}                           & 0.05          & \redcross                                        & \redcross                           \\
DVSMOTION20      & \redcross                                    & \textbf{high}                           & 0.1           & \redcross                                        & \redcross                           \\
MVSEC            & (\redcross)                                  & medium/low                                  & 0.1           & \greencheck                                        & \greencheck                           \\
\textbf{Ours}    & \greencheck                                    & \textbf{high}                           & \textbf{0.3}  & \greencheck                                        & \greencheck                           \\ \bottomrule
\end{tabular}

    \caption{Comparison of existing event camera datasets for optical flow estimation.}
    \label{tab:relw_datasets}
\end{table*}

To the best of our knowledge, there are currently only three event camera datasets that contain optical flow ground truth data.
Table \ref{tab:relw_datasets} summarizes and relates them to our novel proposed dataset.

DVSFLOW16 \cite{rueckauer2016evaluation} and DVSMOTION20 \cite{almatrafi2020distance} are two small datasets designed for evaluating event-based optical flow algorithms for rotational camera motion in indoor scenes. Due to their small size and restricted camera motion, they are not well suited for training and evaluation of neural network based approaches.

More closely related to our work is the MVSEC dataset \cite{Zhu18rss}.
It contains 5 driving sequences and 4 indoor sequences in the same environment.
Groundtruth optical flow data is computed from lidar odometry and mapping for the outdoor scenes.
However, there are three major limitations of this dataset.
First, the ground truth quality is impaired by the lack of proper handling of occlusions and moving objects.
Second, the resolution of the event cameras is low with only 0.1 Megapixels leading to small pixel displacements. We analyze this aspect in detail in section \ref{sec:dataset}.
Finally, MVSEC does not introduce a clear train and test split \cite{Zhu18rss}. For example, the authors evaluate their proposed method on the outdoor day 1 sequence but also train on this sequence to evaluate on the indoor sequences.
We address all these shortcomings by proposing a novel dataset free of ambiguities from occlusions and artifacts from moving objects, 3 times higher camera resolution, and a well-defined training and test split with established metrics to foster reproducibility.

\section{Approach}
Our approach takes inspiration from recent work on two-frame optical flow estimation \cite{teed20eccv}.
However, instead of images, we use consecutive packets of events, which we convert to tensor-like event representations that are compatible with standard CNNs.
Two CNN encoders with shared weights create a feature embedding to compute an all-pairs correlation volume at $1/8$-th of the original resolution.
The same encoder architecture, but with a different set of weights, is used to encode context features for further processing.
Both outputs are used in an iteration scheme to extract per pixel flow, as proposed in \cite{teed20eccv}. 
We then extend this formulation to a temporally recurrent architecture, that processes sequences of event packets, by leveraging differentiable warmstarting of optical flow (Sec. \ref{sec:sequential}). 

\subsection{Problem Definition}
Dense optical flow estimation is commonly defined as the task of finding pixel correspondences between two images $I_i$ and $I_j$ \cite{Menze2015CVPR, Butler:ECCV:2012} denoted as flow $F_{i\to j}$.
For event cameras, we are interested in the analog setting of finding pixel correspondence $F_{t_i\to t_j}$ between two timestamps $t_i$ and $t_j$ given events in between both timestamps.
In other words, our setting is strictly sequential in contrast to image-based optical flow estimation.
For the sake of concise notation, the remaining text refers to $F_{t_i\to t_j}$ as $F_{i\to j}$.

Different from frame-based cameras that acquire images at a specific rate, event cameras respond to brightness changes in the scene asynchronously and independently for each pixel.
They do this by maintaining a per-pixel reference log intensity. As soon as a sufficient deviation from this memorized value is detected, an event is triggered and the memorized value is updated.
An event $e_k(t) = (x_k, y_k, t, p_k)$ is a tuple containing information about its pixel location $(x_k, y_k)$, the exact trigger time $t$ as well as the 1-bit polarity $p_k$ that indicates whether the brightness change was positive or negative.
We define a sequence of events between time $t_i$ and $t_j$ as ${\cal E}(t_i, t_j) := \langle e(t) | t \in [t_i, t_j]\rangle$.

\subsection{RAFT for Event Data}\label{sec:baseline}
Previous approaches for event-based optical flow estimation with neural networks build an input representation given event data between $t_i$ and $t_{i+1} = t_i + \Delta t$ and directly regresses optical flow $F_{i\to i+1}$ \cite{Zhu18rss, Zhu19cvpr, paredes2020back}. Instead, we propose to regress $F_{i\to i+1}$ from \emph{two} short consecutive event sequences ${\cal E}(t_{i-1}, t_i)$ and ${\cal E}(t_{i}, t_{i+1})$.

\begin{figure*}[ht]
    \centering
    \includegraphics[width=0.95\linewidth]{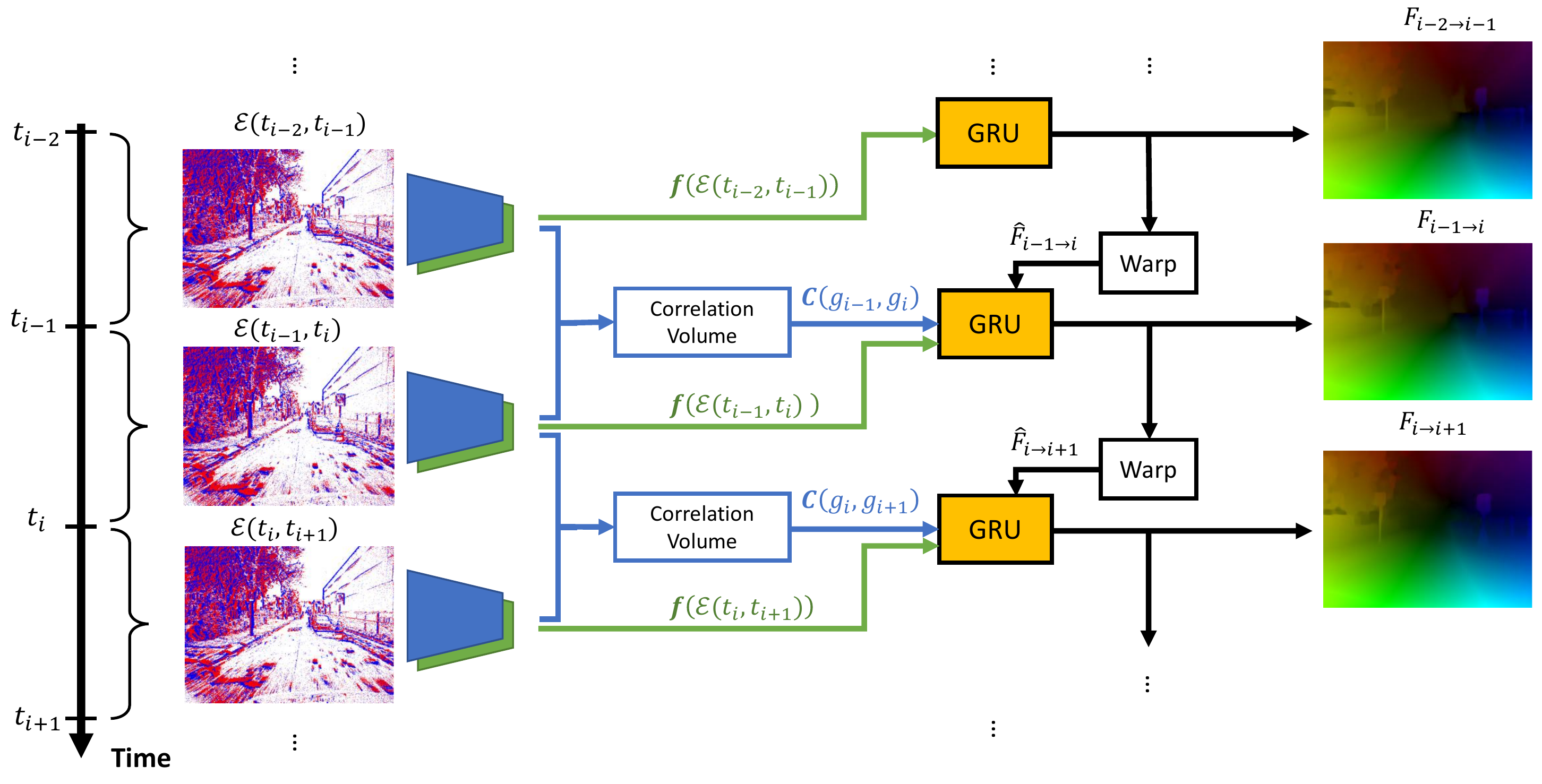}
    \caption{Our proposed recurrent approach for event-based optical flow estimation. The architecture computes a cost volume $C$ from two subsequent event representations. Another context encoder $f$ intializes the hidden state of the GRU update unit and provides additional semantic priors. Our proposed warping module fuses different timestamps together and facilitates propagation of information through time.}
    \label{fig:network}
\end{figure*}

Our baseline approach builds on the RAFT architecture \cite{teed20eccv}. We choose this architecture due to its extensibility to sequential data that we describe in detail in section \ref{sec:sequential} and visualize some components in figure \ref{fig:network}.
RAFT uses three feature embeddings to compute optical flow.
First, two feature embeddings are created with a shared encoder to compute a cost volume, which is illustrated in blue in figure \ref{fig:network}.
Second, a context encoder with a different set of weights creates features, shown in green in figure \ref{fig:network}, that are directly used by a GRU update block \cite{Cho14GRU}.

To adopt this idea for event data, we need to build an event representation that is compatible with CNN layers. Our idea is to split a short sequence of event data into two subsequences.
More specifically, we select an event sequence ${\cal E}_i = {\cal E}(t_{i-1}, t_i)$ where $t_{i-1} = t_i - \Delta t$ and a second event sequence ${\cal E}_{i+1}={\cal E}(t_{i}, t_{i+1})$.
For each event sequence, we create a volumetric voxel grid representation of event data \cite{Zhu19cvpr}.
It discretizes the time dimension but is able to retain most of the temporal aspect of events through bilinear interpolation in time.
The feature correlation encoder produces a feature embedding from ${\cal E}_i$ and ${\cal E}_{i+1}$ while the context encoder only receives the event representation of ${\cal E}_{i+1}$.
The reason for this is that events between $t_i$ to $t_{i+1}$ contain more accurate contextual information for the prediction of $F_{i\to i+1}$ than events before time $t_i$.

\subsection{Differentiable Warm-Starting of Optical Flow}\label{sec:sequential}
This section elaborates on a method to incorporate flow priors from previous time steps not only in the evaluation but also in the training of the model.

By default, the RAFT architecture initializes the flow estimate $F_{i\to {i+1}}$ with 0.
Given sequential data, it is also possible to initialize the optical flow from the last prediction \cite{teed20eccv}.
Different from prior work, we propose to implement this initialization step in a differentiable manner and incorporate the warm-starting module in the training. By doing this, the neural network can incorporate temporal priors and backpropagate gradients through time.
The proposed warm-starting module does not introduce any additional parameters and can be added to already trained networks for further finetuning. 
The overall approach is visualized in figure \ref{fig:network}.

Our goal is to reuse the previous flow prediction $F_{i-1\to i}$ to initialize the iterative update module that eventually estimates $F_{i\to i+1}$.
However, simply copying the previous flow estimate to initialize the current one would not take into account that the source pixel location changed according to the flow itself. Instead, the flow has to be forward warped to the next timestamp.
We achieve this through forward warping of the flow predictions. In forward warping, multiple source pixels can map to the same target pixel leading to ambiguities, which is why we perform average splatting in the target pixel space.
We do this by normalizing the warped flow with the weights derived from the bilinear kernel.
In summary, the flow $F_{{i}\to {i+1}}$ is initialized with the previous flow prediction $F_{{i-1}\to i}$ in the following way:

\begin{align}
    g(x_i) &= x_i + F_{i \to {i+1}}(x_i)\\
    F_{i \to {i+1}}(x) &= \frac{\sum_{\forall x_{i-1}} k_b(x-g(x_{i-1})) F_{{i-1} \to {i}}(x_{i-1})}{\sum_{\forall x_{i-1}} k_b(x-g(x_{i-1}))}
\end{align}

where
\begin{align}
    k_b(a) = \text{max}\{0, 1-|a_x|\}\cdot\text{max}\{0, 1-|a_y|\}
\end{align}
is the bilinear interpolation kernel. These operations are subdifferentiable and can directly be implemented in deep learning frameworks such as PyTorch.

\subsection{Supervision}
We supervise all methods on the $L_1$ distance between optical flow prediction and groundtruth.
The loss function for our method is defined as
\begin{align}
    {\cal L} = \sum_{i = 1}^{N_i}\sum_{k = 1}^{N_k}\gamma^{N_k - k}||F_{i\to i+1}^\text{gt} - F_{i\to i+1}^k||_1\label{eq:supervision}
\end{align}
where $N_i$ is the sequence length, $N_k$ the number of iteration steps per timestep, $F_{i\to i+1}^{\text{gt}}$ the groundtruth for time $i$, and $F_{i\to i+1}^k$ the prediction at the $k$th iteration at time $i$. We set $\gamma = 0.8$. 
\section{\dsecflow{} Dataset}\label{sec:dataset}

This section introduces our proposed optical flow dataset that overcomes the major shortcomings of the MVSEC optical flow dataset \cite{Zhu18rss}. Our proposed dataset extends the publicly available DSEC dataset \cite{gehrig2021dsec} which contains disparity ground truth obtained from a rotating Lidar.
It features VGA resolution event cameras which have 3.4 times the number of pixels compared to the Davis346 used in the MVSEC dataset.
Another advantage of the DSEC dataset is that the disparity maps are cleaned from occlusions, and Lidar points on moving objects have been removed with a stereo matching filtering technique. These aspects build a foundation for our extension of this dataset to optical flow ground truth.

\begin{figure*}[ht]
    \centering
    \newcommand{\localsize}{0.163}
    \includegraphics[width=\localsize\linewidth]{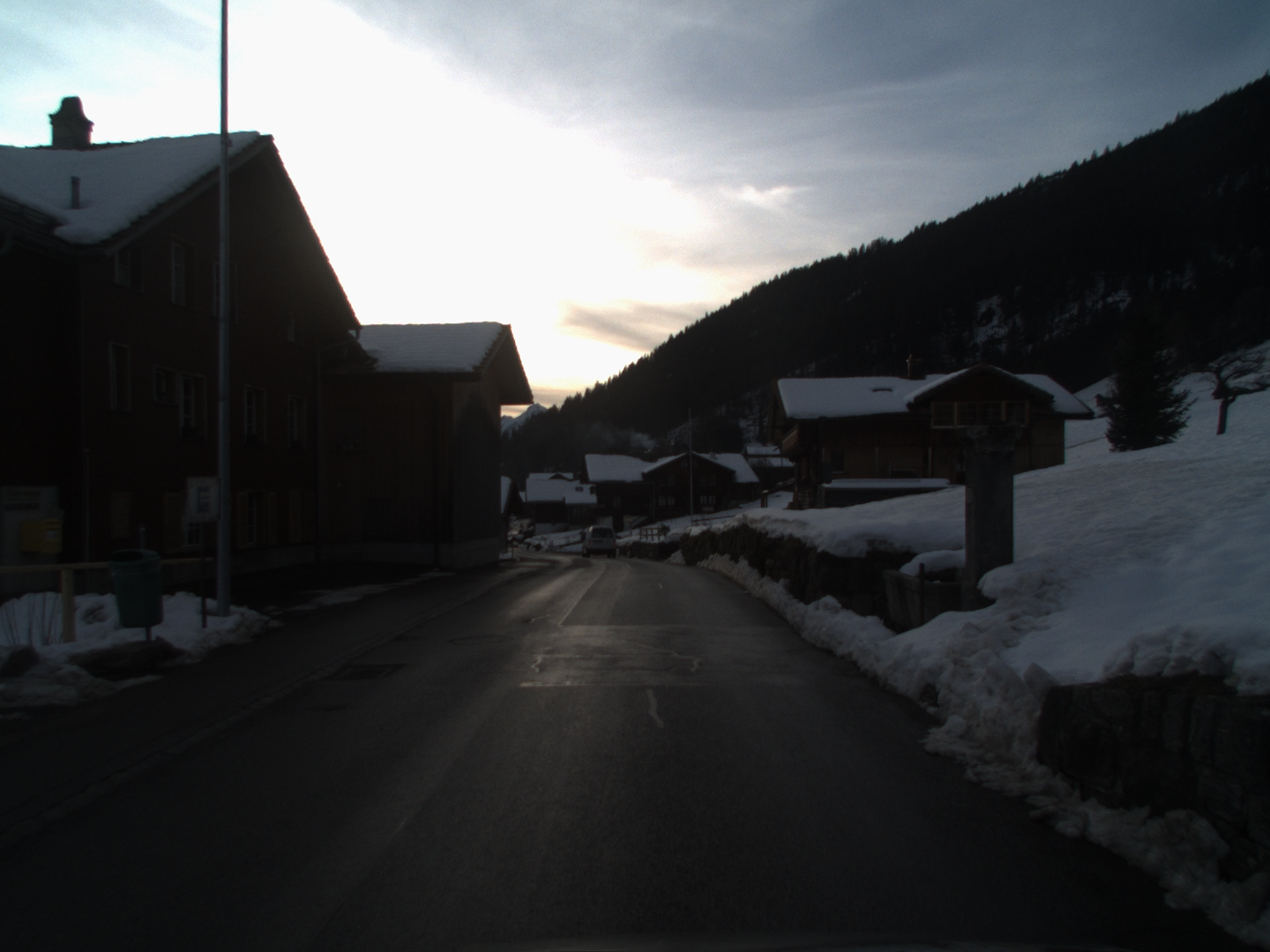}
    \includegraphics[width=\localsize\linewidth]{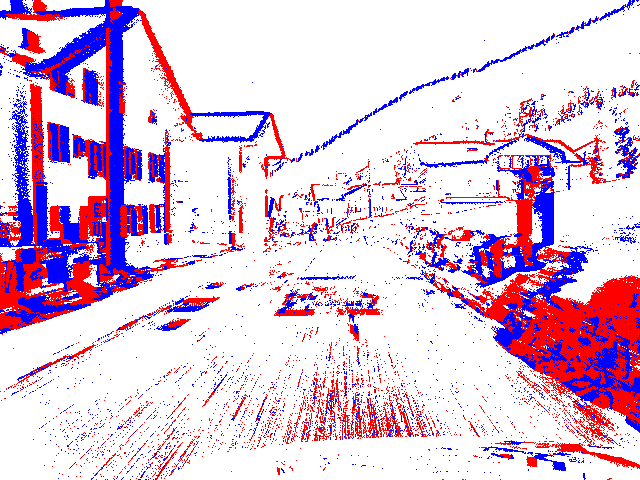}
    \includegraphics[width=\localsize\linewidth]{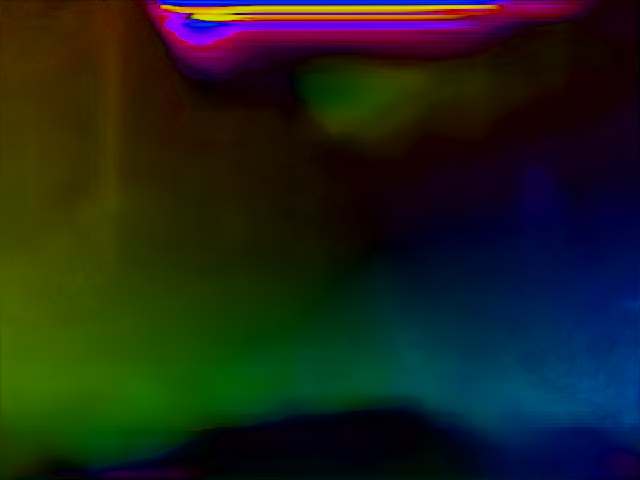}
    \includegraphics[width=\localsize\linewidth]{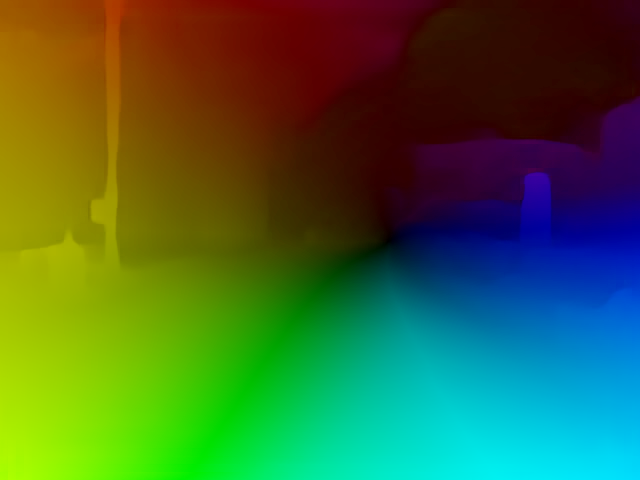}
    \includegraphics[width=\localsize\linewidth]{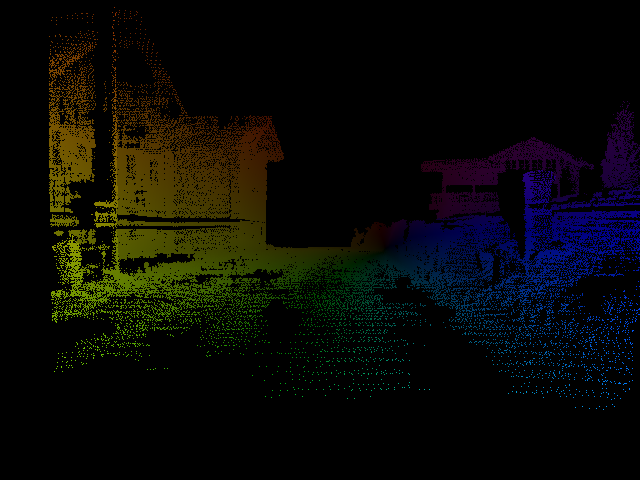}\\
    \includegraphics[width=\localsize\linewidth]{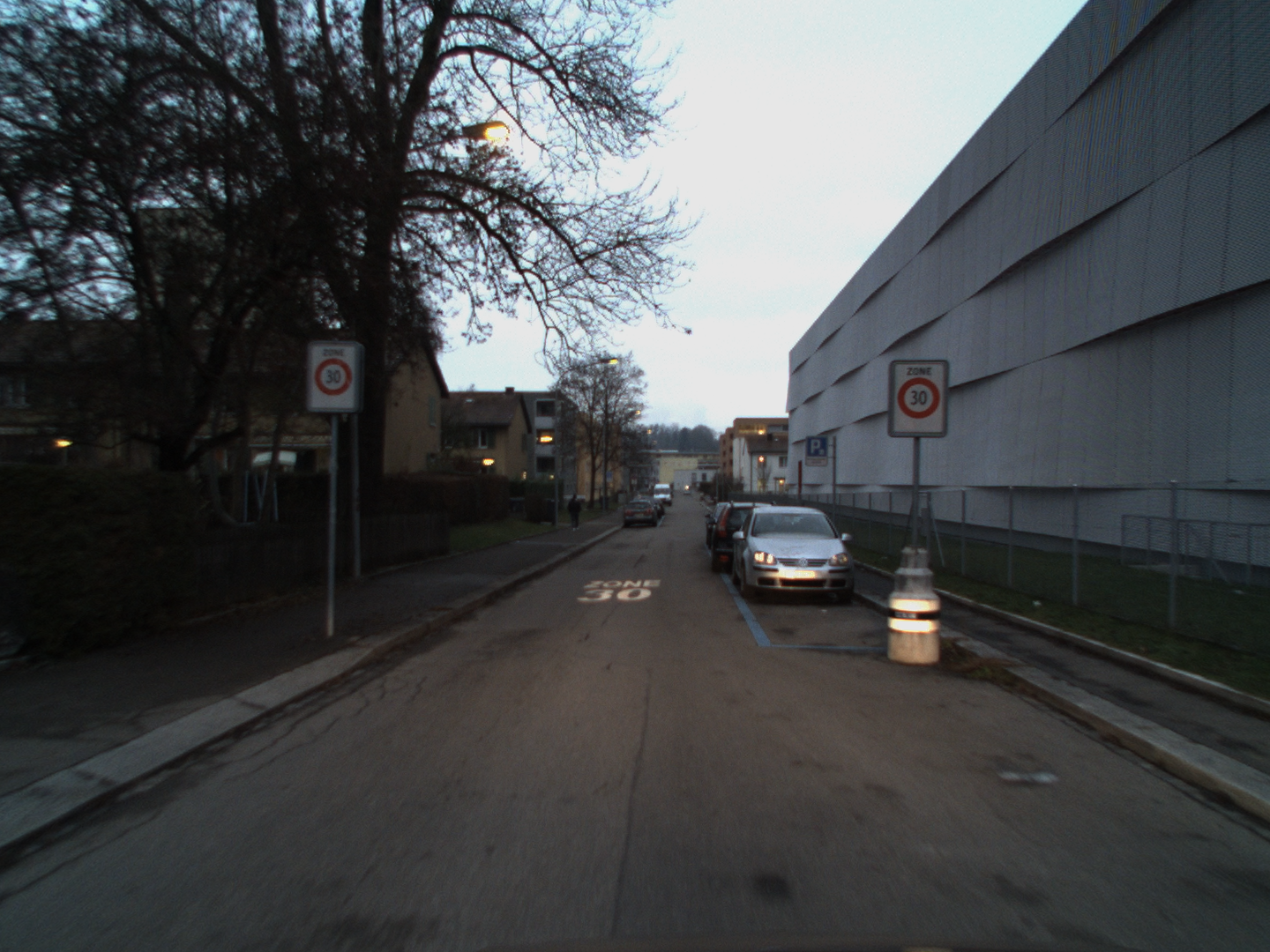}
    \includegraphics[width=\localsize\linewidth]{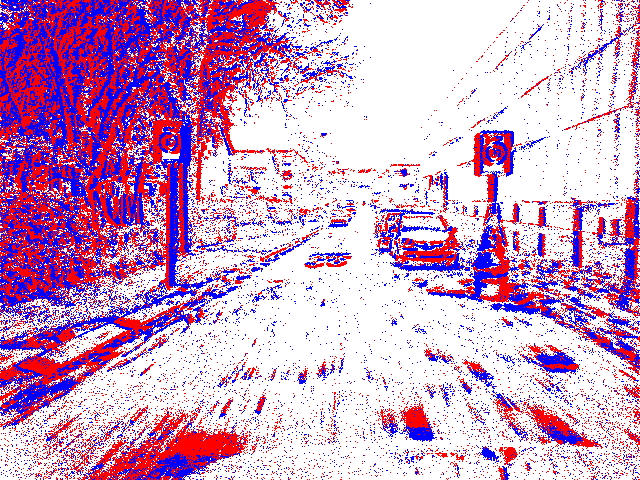}
    \includegraphics[width=\localsize\linewidth]{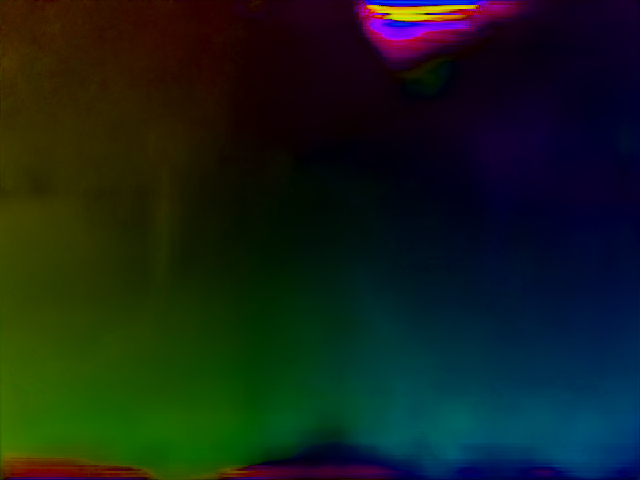}
    \includegraphics[width=\localsize\linewidth]{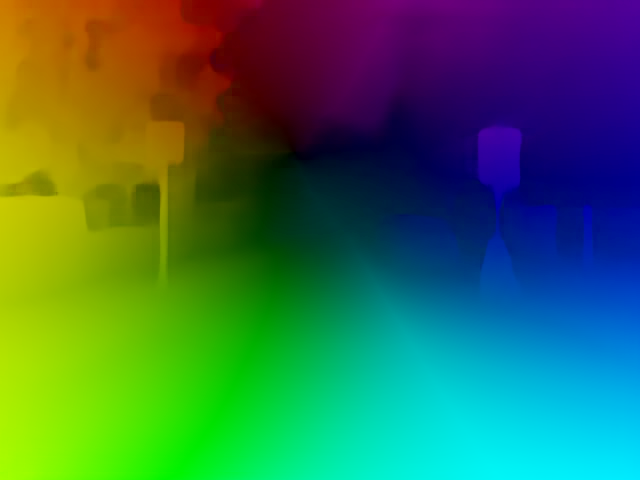}
    \includegraphics[width=\localsize\linewidth]{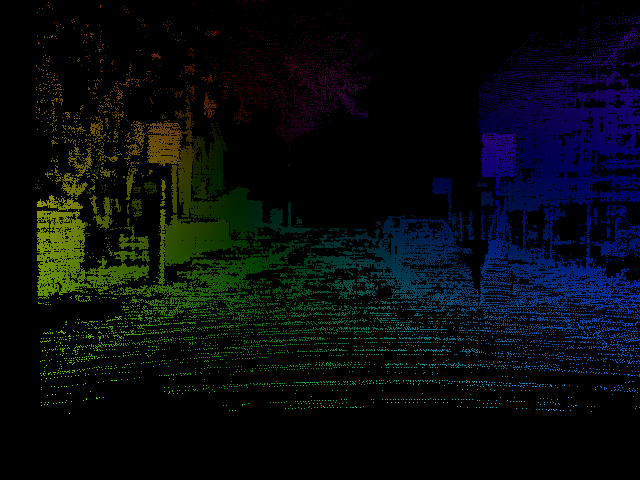}\\
    \includegraphics[width=\localsize\linewidth]{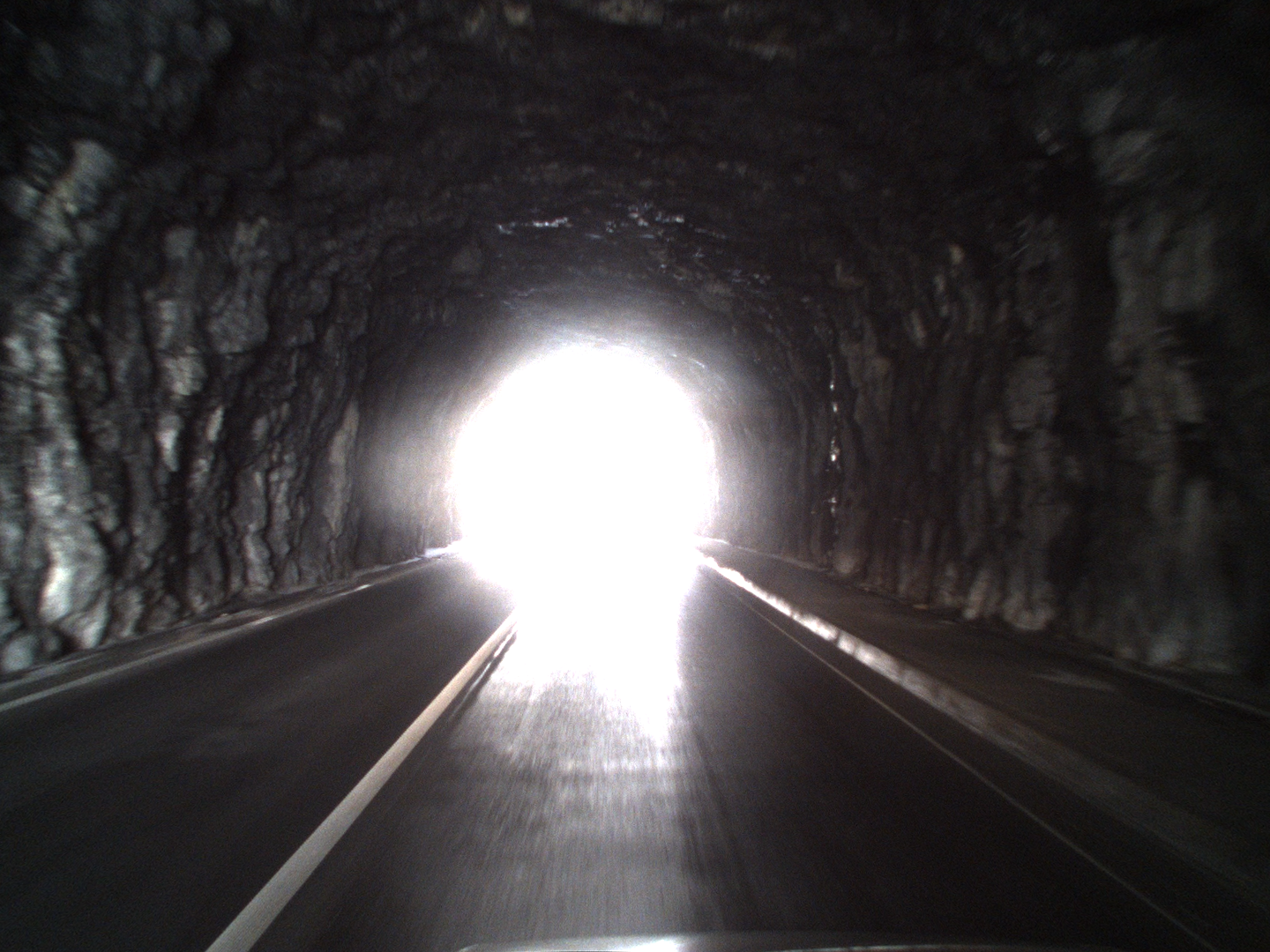}
    \includegraphics[width=\localsize\linewidth]{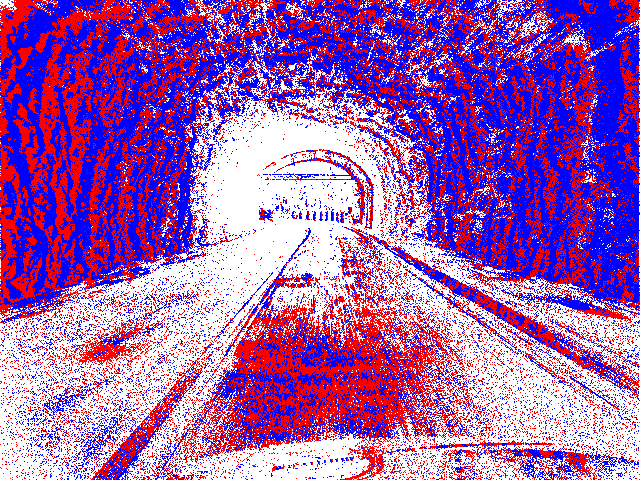}
    \includegraphics[width=\localsize\linewidth]{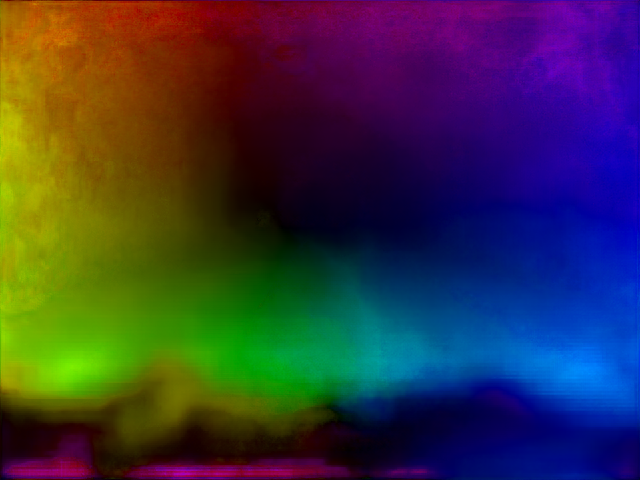}
    \includegraphics[width=\localsize\linewidth]{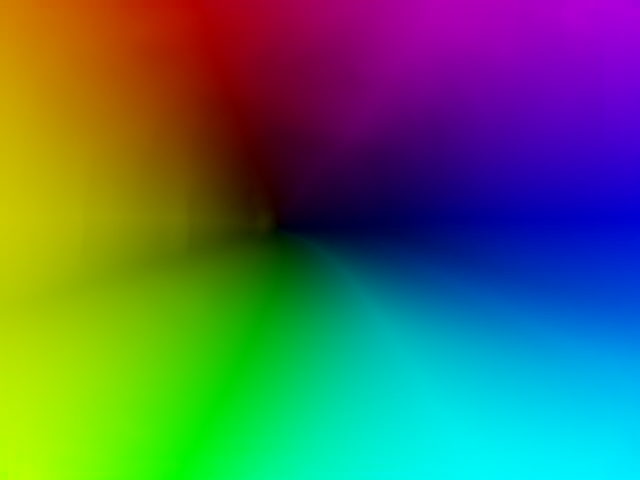}
    \includegraphics[width=\localsize\linewidth]{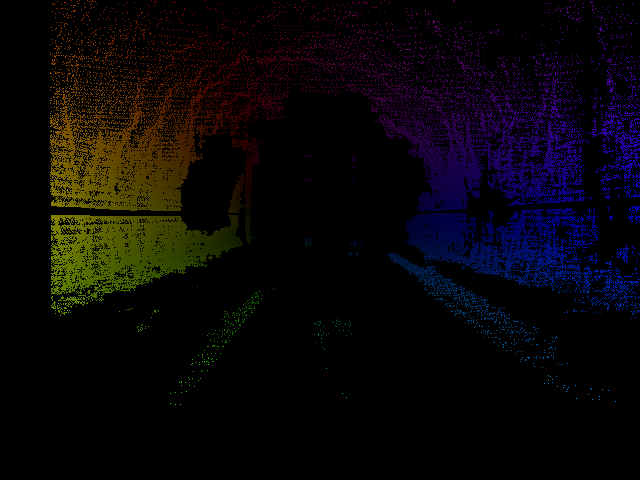}\\
    \includegraphics[width=\localsize\linewidth]{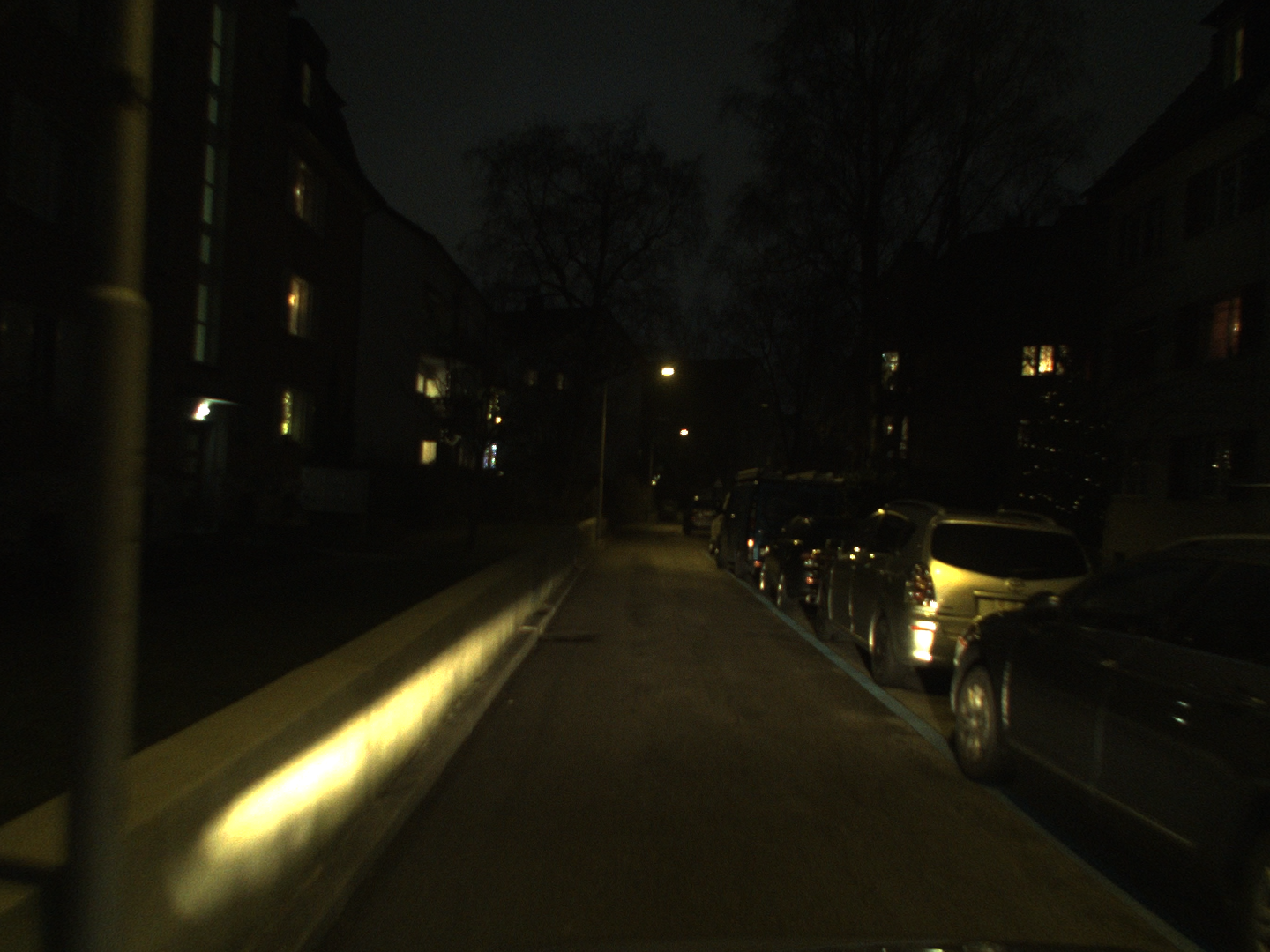}
    \includegraphics[width=\localsize\linewidth]{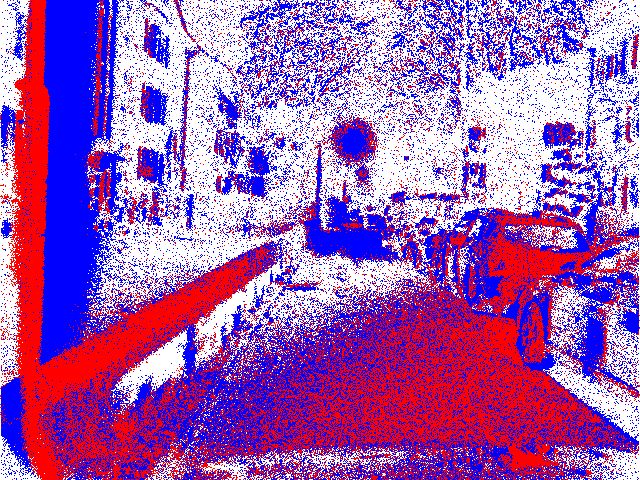}
    \includegraphics[width=\localsize\linewidth]{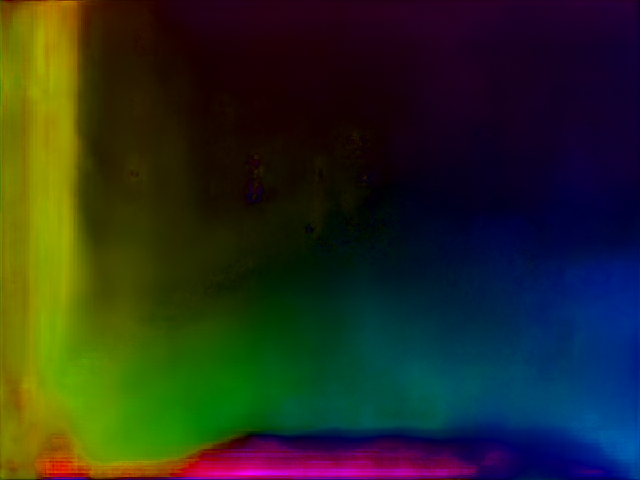}
    \includegraphics[width=\localsize\linewidth]{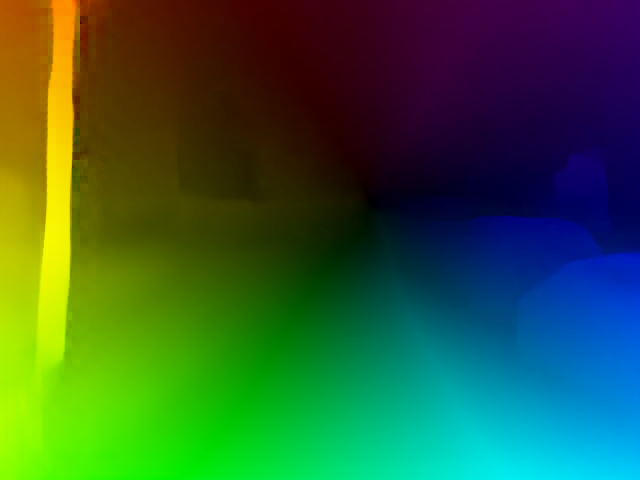}
    \includegraphics[width=\localsize\linewidth]{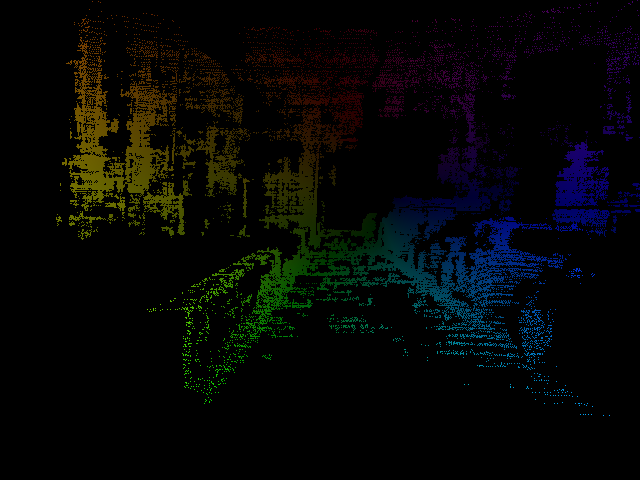}
    \caption{Qualitative examples of optical flow predictions on the \dsecflow{} test set. Best viewed in PDF format. From left to right: Image for visualization only, events visualized as image, EV-Flownet, our approach, and the ground truth. While EV-FlowNet does produce very noisy optical flow, our proposed approach is able to produce very sharp boundaries consistently.}
    \label{fig:def_qual}
\end{figure*}

We utilize the intrinsic and extrinsic calibration data for each sequence to reproject the disparity map into 3D. We then use the odometry ground truth from the dataset to estimate the forward and backward optical flow at 10 Hz assuming that the scene is static.
However, we noticed that, even though most lidar points on dynamic objects have been removed via the proposed filtering technique \cite{gehrig2021dsec}, some points remain on the dynamic objects.
To ensure accurate optical flow ground truth, we manually inspect the whole dataset and extract all parts of sequences that contain static environments.
The resulting dataset contains 7800 training samples and 2100 test samples in 24 sequences at day and night.
Figure \ref{fig:def_qual} shows examples of optical flow predictions on this dataset.

An important aspect that indicates the difficulty of an optical flow dataset is the distribution of pixel displacement magnitudes.
Typically, smaller pixel displacements are easier to predict accurately, in absolute error, than large displacements \cite{Butler:ECCV:2012}. On the other hand, estimating long-range correspondences has proven to be key for accurate, drift-free optical flow estimation for high-speed video \cite{Janai2017CVPR}. Consequently, it is important to devise methods and datasets that address both small and large displacements.

\begin{figure}[ht]
    \centering
    \includegraphics[width=0.8\linewidth]{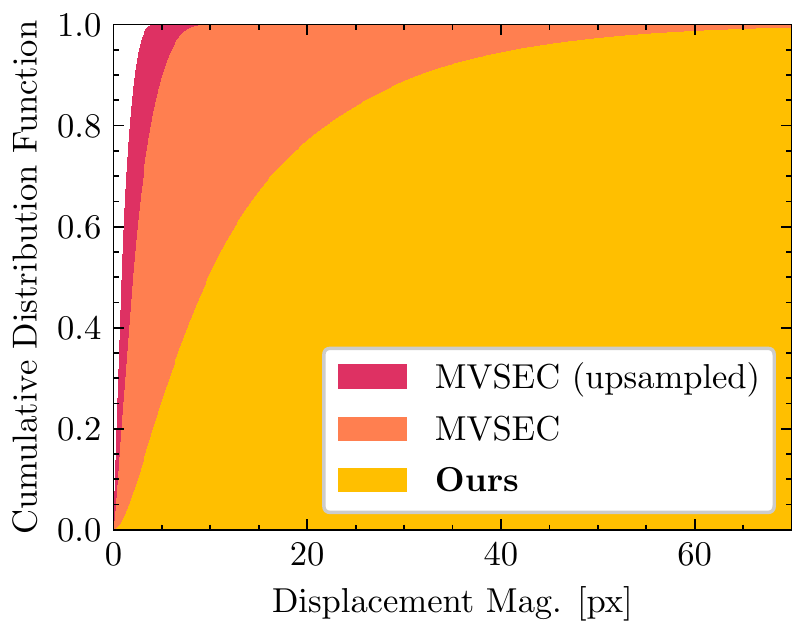}
    \caption{Cumulative Distribution of flow magnitudes normalized by the number of pixels in image width. Our dataset consists of significantly higher optical flow than the MVSEC variants. ``\emph{upsampled}'' indicates that the dataset ground truth has been temporally upsampled according to Zhu et. al \cite{Zhu18rss}}
    \label{fig:flow_hist}
\end{figure}

Figure \ref{fig:flow_hist} analyzes displacement distribution of our dataset and compares the ground truth flow magnitudes of {\dsecflow} with MVSEC.
Optical flow ground truth for MVSEC is originally provided at 20 Hz.
However, following initial evaluation of Zhu et al. \cite{Zhu18rss}, recent work also evaluates on temporally upsampled ground truth at frame rate \cite{lee2020spike,Zhu19cvpr, Ye21iros} at approximately 45 Hz.

Figure \ref{fig:flow_hist} also shows that our dataset consists of drastically higher displacement fields than MVSEC. 80 \% of optical flow magnitudes on MVSEC are lower than 4 pixels.
For the upsampled version this value decreases to mere 1.7 pixels.
In stark contrast, {\dsecflow} reaches 22 pixel flow magnitude for the 80 \% percentile.

For further analysis, we also normalize the flow magnitudes by the image width. The 80 \% percentile for our dataset is 3.4 \% while MVSEC reaches 1.3 \% and 0.5 \% respectively. This indicates that {\dsecflow} is more challenging even when normalized with respect to the resolution of the camera.

\subsection{Evaluation Procedure}
Our test set consists of approximately 2100 ground truth flow maps in seven different sequences. We evaluate optical flow predictions based on the end-point-error (EPE) and also compute the $N$-pixel errors ($N$PE) that indicate the percentage of flow errors higher than $N$ pixels in magnitude. We compute $N$PE for $N$ equals 3, 2 and 1 for coarse to fine analysis of flow outliers, but consider the EPE as the default metric since it provides the most accurate measure of flow accuracy.

\section{Experiments}\label{sec:experiments}

We evaluate our method on MVSEC \cite{Zhu18rss} and the proposed dataset \dsecflow{}.
The published baselines are exclusively self-supervised methods that have been trained and evaluated on the MVSEC dataset.
Since we are interested in understanding the full potential of existing approaches in comparison to our method, we retrain the EV-FlowNet \cite{Zhu18rss, Zhu19cvpr} architecture on both MVSEC and \dsecflow{} with direct supervision from scratch. Note that the EV-FlowNet architecture, which was also used in \cite{Zhu19cvpr}, achieved the lowest EPE of all baselines in the sparse evaluation summarized in table \ref{tab:mvsec_45hz}.\footnote{ECN \cite{Ye21iros} computes static optical flow only indirectly via depth and ego-motion. Therefore, the comparison is not fair.}
We optimize the learning rate for each method separately and perform horizonal flipping and random cropping for data augmention on all datasets.
We also set the number of iterations per timestep, $N_k$ in equation \ref{eq:supervision}, to 12 for training and evaluation on both datasets.

\subsection{MVSEC}

\begin{table*}[ht]
    \centering
    \begin{tabular}{@{}llllllll@{}}
\toprule
                     &           & Dense         &              &     & Sparse        &              &     \\
                     & Loss Type & EPE           & 1PE          & 3PE & EPE           & 1PE          & 3PE \\ \midrule
Back to Event Basics \cite{paredes2020back} & E         & -             & -            & -   & 0.92          & -            & 5.4 \\
Spike-FlowNet \cite{lee2020spike}       & I         & -             & -            & -   & 0.84          & -            & -   \\
ECN * \cite{Ye21iros}               & E         & \textit{0.35}          & -            & 0.0 & \textit{0.30}          & -            & 0.0 \\
EV-FlowNet \cite{Zhu18rss}           & I         & -             & -            & -   & 0.49          & -            & 0.2 \\
Zhu et. al \cite{Zhu19cvpr}          & E         & -             & -            & -   & 0.32          & -            & 0.0 \\ \midrule
EV-FlowNet \cite{Zhu18rss}          & S         & \textit{0.35}          & \textit{3.7}          & 0.0 & 0.31          & \textbf{1.6} & 0.0 \\
\textbf{Ours}        & S         & \textbf{0.27} & \textbf{1.7} & 0.0 & \textbf{0.24} & \textit{1.7} & 0.0 \\ \bottomrule
\end{tabular}

    \caption{MVSEC results at temporally upsampled optical flow ground truth at around 45 Hz. Supervision type E refers to event-based loss, I to image-based (photometric) loss, and S to supervised learning. $*$ indicates that this method predicts optical flow via depth and ego-motion. Best performance in bold and second best in italic. We do not highlight the 3PE as it is zero for almost all methods.}
    \label{tab:mvsec_45hz}
\end{table*}

\begin{table}[ht]
    \centering
    \resizebox{\columnwidth}{!}{
\begin{tabular}{@{}lllllll@{}}
\toprule
              & Dense         &              &               & Sparse        &              &               \\ \midrule
              & EPE           & 1PE          & 3PE           & EPE           & 1PE          & 3PE           \\ \midrule
EV-FlowNet \cite{Zhu18rss}    & 0.61          & 15.6         & 0.45          & 0.53          & 10.7         & 0.60          \\
\textbf{Ours} & \textbf{0.47} & \textbf{9.2} & \textbf{0.24} & \textbf{0.46} & \textbf{7.5} & \textbf{0.49} \\ \bottomrule
\end{tabular}
}
    \caption{MVSEC results at original ground truth rate of 20 Hz. Best performance in bold.}
    \label{tab:mvsec_20hz}
\end{table}

On MVSEC we train our approach with differentiable warm-starting from scratch with sequence length 2 on the outdoor day 2 sequence. After 10 epochs, we increase the sequence length to 5.
We follow prior work and train, and evaluate on the temporally upsampled ground truth. The upsampled ground truth is post-processed to the frame-rate of the camera frames approximately at 45 Hz.
Additionally, we also train, and evaluate EV-FlowNet and our method on the provided ground truth without any modifications to render the task more challenging.

Note that we also report 3-pixel errors on MVSEC to stay consistent with prior work in table \ref{tab:mvsec_45hz}, although 3PE is typically zero.

\paragraph{Sparse vs. Dense} Prior work often masks the optical flow predictions to the pixel locations where events triggered \cite{Zhu18rss, Zhu19cvpr,lee2020spike, paredes2020back}. Since not all methods are open source, we compare against our approaches and the supervised baseline at upsampled ground truth. Table \ref{tab:mvsec_45hz} shows that the metrics are consistently lower when the evaluation is performed with masked ground truth. The sparse EPE is 11\% lower compared to the dense EPE. Hence, dense optical flow prediction is a more difficult task on MVSEC. With a dense EPE of 0.27 and sparse EPE of 0.24, our approach outperforms the sparse and dense baselines and methods proposed in the literature. 

\paragraph{Supervised vs. Self-Supervised} Our approach and the retrained EV-FlowNet baseline are the only supervised networks in the literature on this dataset. Regardeless of sparse or dense evaluation, supervised methods generally outperform self-supervised methods. The worse performing supervised baseline, EV-FlowNet, achieves a sparse EPE of 0.31 while Zhu et al. \cite{Zhu19cvpr} reach 0.32 EPE. We hypothesize that the EPE error advantage of supervised learning methods would be significantly higher if MVSEC ground truth optical flow would be of higher quality.

\paragraph{Temporally Upsampled vs Original Groundtruth Rate} Table \ref{tab:mvsec_20hz} illustrates the performance of supervised EV-FlowNet and our method on MVSEC without ground truth upsampling. The direct comparison of our method and the supervised EV-FlowNet on both MVSEC versions reveils that the EPE of our approach is 23 \% lower in both cases. Hence, the original (20 Hz) optical flow of MVSEC does not yet reveil additional insights with respect to the performance of estimating larger displacement fields. The experiments of the next section \ref{exp:def} addresses this challenge. Nonetheless, the 1-pixel error increases substantially for both methods. This is, of course, expected because the optical flow magnitudes increase due to the larger time windows.

\subsection{\dsecflow}\label{exp:def}
On the \dsecflow{} dataset we perform horizontal flipping and random cropping to 288 pixels in height and 384 pixels in width. We train our non-recurrent approach  with a learning rate of $10^{-4}$ with Adam for 40 epochs until convergence.
Then, we introduce the differentiable warm-starting with a sequence length of 3 and train for another 6 epochs with random cropping.
Finally, we reduce the learning rate by a factor of 10 and fine-tune on the full resolution for evaluation.
For a fair comparison, we apply the same fine-tuning schedule to the non-recurrent architecture as well as the baseline model.

Table \ref{tab:dsec_metrics} compares our approach against the baseline method. Our method achieves an EPE of 0.79 which is 2.9 times lower than EV-FlowNet with an EPE of 2.32. To our surprise, EV-FlowNet does not perform comparably in \dsecflow{}. This is different to the MVSEC experiments where the performance difference between our method and EV-FlowNet was in a comparable range. A possible explanation is that EV-FlowNet was designed on MVSEC to work for small pixel displacements.

\begin{table}[ht]
    \centering
    \begin{tabular}{@{}lllll@{}}
\toprule
              & EPE           & 1PE           & 2PE          & 3PE          \\ \midrule
EV-FlowNet \cite{Zhu18rss}    & 2.32          & 55.4          & 29.8         & 18.6         \\
\textbf{Ours} & \textbf{0.79} & \textbf{12.5} & \textbf{4.7} & \textbf{2.7} \\ \bottomrule
\end{tabular}

    \caption{Evaluation on the \dsecflow{} dataset. Best performance in bold.}
    \label{tab:dsec_metrics}
\end{table}

\subsection{Ablation Study}\label{exp:ablation}
This section introduces the ablation studies that highlight the significance of different modules.
We perform these ablation studies both on the \dsecflow{} dataset and on the MVSEC dataset.
Table \ref{tab:ablation} summarizes the results of the ablation study that focuses on warm-starting strategies in combination with different evaluation and training combinations.

\begin{table*}[ht]
    \centering
    \begin{tabular}{@{}lllllll@{}}
    \toprule
                                     & \multicolumn{4}{l}{\dsecflow}                               & \multicolumn{2}{l}{MVSEC}    \\ \cmidrule(l){2-7} 
                                     & EPE           & 1PE           & 2PE          & 3PE          & EPE           & 1PE          \\ \midrule
    Ours w/o WS                      & 0.82          & 13.5          & 5.2          & 3.0          & 0.53          & 12.2         \\
    Ours eval WS                     & 0.88          & 14.1          & 5.6          & 3.3          & 0.51          & 10.9         \\
    Ours w/o WS: 100 iters           & 0.87          & 14.9          & 5.9          & 3.4          & 0.60          & 16.3         \\
    Ours train \& eval WS: 100 iters & 0.88          & 15.1          & 6.0          & 3.5          & 0.49          & 10.0         \\
    Ours train \& eval WS            & \textbf{0.79} & \textbf{12.7} & \textbf{4.7} & \textbf{2.7} & \textbf{0.47} & \textbf{9.2} \\ \bottomrule
\end{tabular}

    \caption{Ablation study on \dsecflow{} and MVSEC. Best performance in bold. \emph{WS} refers to warm-starting optical flow from the previous estimated optical flow. \emph{eval WS} means that warm-starting is only used at inference time without training with warm-starting beforehand.}
    \label{tab:ablation}
\end{table*}
The baseline method (Ours w/o WS) is the proposed non-recurrent architecture applied to event data and does not incorporate any warm-starting in training or evaluation. 

\paragraph{\dsecflow{} Dataset} On the \dsecflow{} dataset, the baseline approach performs better compared to evaluating the same method with warm-starting. That is without incorporating warm-starting in the training. Surprisingly, it also outperforms both evaluation protocols that involve 100 iterations per timestep, regardless whether we train and evaluate with warm-starting or not. Our proposed approach, which incorporates warm-starting in the training and evaluation, outperforms the ablated methods with 4.0 \% lower EPE of 0.79.

\paragraph{MVSEC Dataset} The results on the MVSEC ablation similarly suggest that training and evaluating with warm-starting is superior to the aforementioned variations. However, differently from \dsecflow{}, purely evaluating with warm-starting (without training) leads to lower errors than not using warm-starting at all. Still, the model that also incorporates training with the warm-starting module achieves 4.2 \% lower EPE than the closest ablated method (training + evaluation + 100 iterations).\\

Overall, the effect of incorporating warm-starting into the fine-tuning procedure improves the performance consistently by a few percent on both datasets. Surprisingly, such gains are not consistent in our experiments if we only add warm-starting at inference time as reported in \cite{teed20eccv}.

\section{Conclusion}

This work proposes an alternative approach to event-based optical flow estimation by adopting design choices from frame-based optical flow estimation and modifying the method to integrate with event-based input. We showed that the proposed method drastically outperforms baseline methods on two datasets, which suggests that the commonly used U-Net architecture in the event-based literature is a suboptimal choice for the task at hand. A key contribution of this work is a novel real-world optical flow dataset that addresses the major shortcomings of existing event-based datasets. We showed that training on this dataset leads to high-quality optical flow predictions that are unattainable with previous event-based vision datasets.

\section{Acknowledgments}
This work was supported by the National Centre of Competence in Research (NCCR) Robotics through the Swiss National Science Foundation (SNSF) and by the European Research Council (ERC) under Grant Agreement 864042 (AGILEFLIGHT)

{\small
\bibliographystyle{ieee_fullname}
\bibliography{all}

\begin{thebibliography}{10}\itemsep=-1pt

\bibitem{almatrafi2020distance}
Mohammed Almatrafi, Raymond Baldwin, Kiyoharu Aizawa, and Keigo Hirakawa.
\newblock Distance surface for event-based optical flow.
\newblock {\em IEEE transactions on pattern analysis and machine intelligence},
  2020.

\bibitem{Bardow16cvpr}
Patrick Bardow, Andrew~J. Davison, and Stefan Leutenegger.
\newblock Simultaneous optical flow and intensity estimation from an event
  camera.
\newblock In {\em {IEEE} Conf. Comput. Vis. Pattern Recog. (CVPR)}, pages
  884--892, 2016.

\bibitem{Benosman14tnnls}
Ryad Benosman, Charles Clercq, Xavier Lagorce, Sio-Hoi Ieng, and Chiara
  Bartolozzi.
\newblock Event-based visual flow.
\newblock {\em {IEEE} Trans. Neural Netw. Learn. Syst.}, 25(2):407--417, 2014.

\bibitem{benosman2012asynchronous}
Ryad Benosman, Sio-Hoi Ieng, Charles Clercq, Chiara Bartolozzi, and Mandyam
  Srinivasan.
\newblock Asynchronous frameless event-based optical flow.
\newblock {\em Neural Networks}, 27:32--37, 2012.

\bibitem{brosch2015event}
Tobias Brosch, Stephan Tschechne, and Heiko Neumann.
\newblock On event-based optical flow detection.
\newblock {\em Frontiers in neuroscience}, 9:137, 2015.

\bibitem{Butler:ECCV:2012}
D.~J. Butler, J. Wulff, G.~B. Stanley, and M.~J. Black.
\newblock A naturalistic open source movie for optical flow evaluation.
\newblock In {A. Fitzgibbon et al. (Eds.)}, editor, {\em European Conf. on
  Computer Vision (ECCV)}, Part IV, LNCS 7577, pages 611--625. Springer-Verlag,
  Oct. 2012.

\bibitem{Cho14GRU}
Kyunghyun Cho, B {van Merrienboer}, Caglar Gulcehre, F Bougares, H Schwenk, and
  Yoshua Bengio.
\newblock Learning phrase representations using rnn encoder-decoder for
  statistical machine translation.
\newblock In {\em Conference on Empirical Methods in Natural Language
  Processing (EMNLP 2014)}, 2014.

\bibitem{Dosovitskiy15iccv}
Alexey Dosovitskiy, Philipp Fischer, Eddy Ilg, Philip H{\"a}usser, Caner
  Haz{\i}rba{\c{s}}, Vladimir Golkov, Patrick van~der Smagt, Daniel Cremers,
  and Thomas Brox.
\newblock Flow{N}et: Learning optical flow with convolutional networks.
\newblock In {\em Int. Conf. Comput. Vis. (ICCV)}, pages 2758--2766, 2015.

\bibitem{Gallego20pami}
Guillermo Gallego, Tobi Delbruck, Garrick Orchard, Chiara Bartolozzi, Brian
  Taba, Andrea Censi, Stefan Leutenegger, Andrew Davison, J{\"o}rg Conradt,
  Kostas Daniilidis, and Davide Scaramuzza.
\newblock Event-based vision: A survey.
\newblock {\em {IEEE} Trans. Pattern Anal. Mach. Intell.}, 2020.

\bibitem{Gehrig19ijcv}
Daniel Gehrig, Henri Rebecq, Guillermo Gallego, and Davide Scaramuzza.
\newblock {EKLT}: Asynchronous photometric feature tracking using events and
  frames.
\newblock {\em Int. J. Comput. Vis.}, 2019.

\bibitem{gehrig2021dsec}
Mathias Gehrig, Willem Aarents, Daniel Gehrig, and Davide Scaramuzza.
\newblock Dsec: A stereo event camera dataset for driving scenarios.
\newblock {\em {IEEE} Robot. Autom. Lett.}, 2021.

\bibitem{Geiger13ijrr}
Andreas Geiger, Philip Lenz, Christoph Stiller, and Raquel Urtasun.
\newblock Vision meets robotics: The {KITTI} dataset.
\newblock {\em Int. J. Robot. Research}, 32(11):1231--1237, 2013.

\bibitem{Horn81ai}
Berthold~K.P. Horn and Brian~G. Schunck.
\newblock Determining optical flow.
\newblock {\em J. Artificial Intell.}, 17(1):185 -- 203, 1981.

\bibitem{hui2018liteflownet}
Tak-Wai Hui, Xiaoou Tang, and Chen~Change Loy.
\newblock Liteflownet: A lightweight convolutional neural network for optical
  flow estimation.
\newblock In {\em {IEEE} Conf. Comput. Vis. Pattern Recog. (CVPR)}, 2018.

\bibitem{hur2019iterative}
Junhwa Hur and Stefan Roth.
\newblock Iterative residual refinement for joint optical flow and occlusion
  estimation.
\newblock In {\em {IEEE} Conf. Comput. Vis. Pattern Recog. (CVPR)}, 2019.

\bibitem{janai2020computer}
Joel Janai, Fatma G{\"u}ney, Aseem Behl, Andreas Geiger, et~al.
\newblock Computer vision for autonomous vehicles: Problems, datasets and state
  of the art.
\newblock {\em Foundations and Trends{\textregistered} in Computer Graphics and
  Vision}, 12(1--3):1--308, 2020.

\bibitem{Janai2017CVPR}
Joel Janai, Fatma Güney, Jonas Wulff, Michael Black, and Andreas Geiger.
\newblock Slow flow: Exploiting high-speed cameras for accurate and diverse
  optical flow reference data.
\newblock In {\em CVPR}, 2017.

\bibitem{kalantari2017deep}
Nima~Khademi Kalantari and Ravi Ramamoorthi.
\newblock Deep high dynamic range imaging of dynamic scenes.
\newblock {\em ACM Trans. Graph.}, 36(4):144--1, 2017.

\bibitem{lee2020spike}
Chankyu Lee, Adarsh~Kumar Kosta, Alex~Zihao Zhu, Kenneth Chaney, Kostas
  Daniilidis, and Kaushik Roy.
\newblock Spike-flownet: event-based optical flow estimation with
  energy-efficient hybrid neural networks.
\newblock In {\em Eur. Conf. Comput. Vis. (ECCV)}, 2020.

\bibitem{Lichtsteiner08ssc}
Patrick Lichtsteiner, Christoph Posch, and Tobi Delbruck.
\newblock {A 128$\times$128 120 dB 15 $\mu$s latency asynchronous temporal
  contrast vision sensor}.
\newblock {\em {IEEE} J. Solid-State Circuits}, 43(2):566--576, 2008.

\bibitem{liu2019selflow}
Pengpeng Liu, Michael Lyu, Irwin King, and Jia Xu.
\newblock Selflow: Self-supervised learning of optical flow.
\newblock In {\em {IEEE} Conf. Comput. Vis. Pattern Recog. (CVPR)}, 2019.

\bibitem{Lucas81ijcai}
Bruce~D. Lucas and Takeo Kanade.
\newblock An iterative image registration technique with an application to
  stereo vision.
\newblock In {\em Int. Joint Conf. Artificial Intell. (IJCAI)}, pages 674--679,
  1981.

\bibitem{Menze2015CVPR}
Moritz Menze and Andreas Geiger.
\newblock Object scene flow for autonomous vehicles.
\newblock In {\em {IEEE} Conf. Comput. Vis. Pattern Recog. (CVPR)}, 2015.

\bibitem{Mueggler15icra}
Elias Mueggler, Christian Forster, Nathan Baumli, Guillermo Gallego, and Davide
  Scaramuzza.
\newblock Lifetime estimation of events from dynamic vision sensors.
\newblock In {\em {IEEE} Int. Conf. Robot. Autom. (ICRA)}, pages 4874--4881,
  2015.

\bibitem{niklaus20cvpr}
Simon Niklaus and Feng Liu.
\newblock Softmax splatting for video frame interpolation.
\newblock In {\em {IEEE} Conf. Comput. Vis. Pattern Recog. (CVPR)}, June 2020.

\bibitem{pan2020single}
Liyuan Pan, Miaomiao Liu, and Richard Hartley.
\newblock Single image optical flow estimation with an event camera.
\newblock {\em {IEEE} Conf. Comput. Vis. Pattern Recog. (CVPR)}, 2020.

\bibitem{paredes2020back}
Federico Paredes-Vall{\'e}s and Guido~CHE de Croon.
\newblock Back to event basics: Self-supervised learning of image
  reconstruction for event cameras via photometric constancy.
\newblock {\em arXiv preprint arXiv:2009.08283}, 2020.

\bibitem{qin2018vins}
Tong Qin, Peiliang Li, and Shaojie Shen.
\newblock Vins-mono: A robust and versatile monocular visual-inertial state
  estimator.
\newblock {\em IEEE Transactions on Robotics}, 34(4):1004--1020, 2018.

\bibitem{ronneberger2015u}
Olaf Ronneberger, Philipp Fischer, and Thomas Brox.
\newblock U-net: Convolutional networks for biomedical image segmentation.
\newblock In {\em International Conference on Medical image computing and
  computer-assisted intervention}, pages 234--241. Springer, 2015.

\bibitem{rueckauer2016evaluation}
Bodo Rueckauer and Tobi Delbruck.
\newblock Evaluation of event-based algorithms for optical flow with
  ground-truth from inertial measurement sensor.
\newblock {\em Frontiers in neuroscience}, 2016.

\bibitem{Karen14NeurIPS}
Karen Simonyan and Andrew Zisserman.
\newblock Two-stream convolutional networks for action recognition in videos.
\newblock In {\em Conf. Neural Inf. Process. Syst. (NeurIPS)}, 2014.

\bibitem{Sun18cvpr}
Deqing Sun, Xiaodong Yang, Ming-Yu Liu, and Jan Kautz.
\newblock {PWC-Net}: {CNNs} for optical flow using pyramid, warping, and cost
  volume.
\newblock In {\em {IEEE} Conf. Comput. Vis. Pattern Recog. (CVPR)}, 2018.

\bibitem{sun2019models}
Deqing Sun, Xiaodong Yang, Ming-Yu Liu, and Jan Kautz.
\newblock Models matter, so does training: An empirical study of cnns for
  optical flow estimation.
\newblock {\em IEEE transactions on pattern analysis and machine intelligence},
  2019.

\bibitem{teed20eccv}
Zachary Teed and Jia Deng.
\newblock Raft: Recurrent all-pairs field transforms for optical flow.
\newblock In {\em Eur. Conf. Comput. Vis. (ECCV)}, 2020.

\bibitem{Xu_2017_CVPR}
Jia Xu, Rene Ranftl, and Vladlen Koltun.
\newblock Accurate optical flow via direct cost volume processing.
\newblock In {\em {IEEE} Conf. Comput. Vis. Pattern Recog. (CVPR)}, July 2017.

\bibitem{yang2019volumetric}
Gengshan Yang and Deva Ramanan.
\newblock Volumetric correspondence networks for optical flow.
\newblock {\em NeurIPS}, 5:12, 2019.

\bibitem{Ye21iros}
C. {Ye}, A. {Mitrokhin}, C. {Fermüller}, J.~A. {Yorke}, and Y. {Aloimonos}.
\newblock Unsupervised learning of dense optical flow, depth and egomotion with
  event-based sensors.
\newblock In {\em 2020 IEEE/RSJ International Conference on Intelligent Robots
  and Systems (IROS)}, 2020.

\bibitem{Ye19arxiv}
Chengxi Ye, Anton Mitrokhin, Chethan Parameshwara, Cornelia Ferm\"uller,
  James~A. Yorke, and Yiannis Aloimonos.
\newblock Unsupervised learning of dense optical flow and depth from sparse
  event data.
\newblock {\em ar{X}iv e-prints}, 2019.

\bibitem{zhao2020maskflownet}
Shengyu Zhao, Yilun Sheng, Yue Dong, Eric~I Chang, Yan Xu, et~al.
\newblock Maskflownet: Asymmetric feature matching with learnable occlusion
  mask.
\newblock In {\em {IEEE} Conf. Comput. Vis. Pattern Recog. (CVPR)}, 2020.

\bibitem{Zhu18rss}
Alex~Zihao Zhu, Liangzhe Yuan, Kenneth Chaney, and Kostas Daniilidis.
\newblock {EV-FlowNet}: Self-supervised optical flow estimation for event-based
  cameras.
\newblock In {\em Robotics: Science and Systems (RSS)}, 2018.

\bibitem{Zhu19cvpr}
Alex~Zihao Zhu, Liangzhe Yuan, Kenneth Chaney, and Kostas Daniilidis.
\newblock Unsupervised event-based learning of optical flow, depth, and
  egomotion.
\newblock In {\em {IEEE} Conf. Comput. Vis. Pattern Recog. (CVPR)}, 2019.

\end{thebibliography}
}

\end{document}